# Investigation of event-based memory surfaces for high-speed tracking, unsupervised feature extraction, and object recognition

Saeed Afshar, Student *Member IEEE*, Gregory Cohen, *Member IEEE*, Tara Julia Hamilton, *Member IEEE*,
Jonathan Tapson, *Member IEEE*, André van Schaik, *Fellow IEEE*.
**Biomedical Engineering and Neuroscience Program,**
**The MARCS Institute for Brain, Behaviour, and Development,**
**Western Sydney University, Australia**
s.afshar@westernsydney.edu.au

*Abstract*— In this paper we compare event-based decaying and time based-decaying memory surfaces for high-speed event-based tracking, feature extraction, and object classification using an event-based camera. The high-speed recognition task involves detecting and classifying model airplanes that are dropped free-hand close to the camera lens so as to generate a challenging dataset exhibiting significant variance in target velocity. This variance motivated the investigation of event-based decaying memory surfaces in comparison to time-based decaying memory surfaces to capture the temporal aspect of the event-based data. These surfaces are then used to perform unsupervised feature extraction, tracking and recognition. In order to generate the memory surfaces, event binning, linearly decaying kernels, and exponentially decaying kernels were investigated with exponentially decaying kernels found to perform best. Event-based decaying memory surfaces were found to outperform time-based decaying memory surfaces in recognition especially when invariance to target velocity was made a requirement. A range of network and receptive field sizes were investigated. The system achieves 98.75% recognition accuracy within 156 milliseconds of an airplane entering the field of view, using only twenty-five event-based feature extracting neurons in series with a linear classifier. By comparing the linear classifier results to an ELM classifier, we find that a small number of event-based feature extractors can effectively project the complex spatio-temporal event patterns of the dataset to an almost linearly separable representation in feature space.

## I. Introduction

The last decade has seen significant development in the field of event-based cameras. Cameras such as the Dynamic Vision Sensor (DVS) [1] and the Asynchronous Time-based Image Sensor (ATIS) [2] attempt to model the operation of the human retina by generating events at each pixel in response to changes in illumination. These cameras have spurred the development of a range of visual processing algorithms to tackle existing problems such as optical flow detection [3], scene stitching [4], tracking [5], motion analysis [6], hand gesture recognition [7], hierarchical feature recognition [8], and unsupervised visual feature extraction and learning [9][10].

More recently, the Hierarchy of Time Surfaces (HOTS) [11] was introduced and forms the basis for the research presented here. HOTS makes use of layers of time-decaying event-surfaces and feature-based clustering, with the features learnt in an unsupervised manner. The HOTS approach processes events in the temporal domain and is functionally similar to the feature extraction layer used in this work. Here we use the ATIS camera to perform a similar unsupervised feature extraction operation in addition to tracking and classification.

## II. Methodology

### A. Generating the Dataset

The system presented in this paper constitutes an event-based and high-speed classification system, and makes use of a real-world task, and its associated dataset, to demonstrate and characterize its performance.

A variety of event-based datasets now exist, such as the N-MNIST and N-Caltech101 datasets [12], MNIST_DVS [13] and the event-based UCF-50 [14]. One common facet of these datasets is that they have been generated under highly constrained conditions, especially with respect to the range of target object velocities. For a static image, event-based cameras only produce data in response to motion and therefore require either the static image or the camera itself to be moving. Therefore, the velocities involved in many of the event-based datasets are strictly controlled. This is often a desirable trait to ensure consistency across all samples, but this constraint is a strongly artificial one. Other event-based datasets, such as the visual navigation dataset found in [15], do not control velocity in the same manner, but represent a fundamentally different task and are therefore not well suited to exploring tracking and feature extraction mechanisms.

The need to explore the effect of variances in velocity is important as these tend to produce significant variance in the spatio-temporal event patterns generated by event-based cameras. This can have a significant impact on the performance of a classifier or tracking algorithm. A primary focus of this work is on the comparison of different event-based processing approaches in the presence of such variance. This required the creation of a new dataset designed to test event-based classification algorithms under conditions that are less constrained and closer to those found in real-world tasks. Instead of converting an existing dataset to an event-based format, a novel classification task was designed around a task that is difficult to tackle with conventional image sensors.

The task centers around the identification of model airplanes as they rapidly pass through the field of view of an ATIS camera. The airplanes were dropped free-hand, and from varying heights and distance from the camera as shown in

Figure **1**(a). Four model airplanes were used, each made from steel and all painted uniform gray as shown in

Figure **1**(b). This served to remove any distinctive textures or marking from the airplanes, thereby increasing the difficulty of the task. The airplanes are models of a Mig-31, an F-117, a Su-24, and a Su-35 with wingspans of 9.1 cm, 7.5 cm, 10.3 cm, and 9.0 cm, respectively.

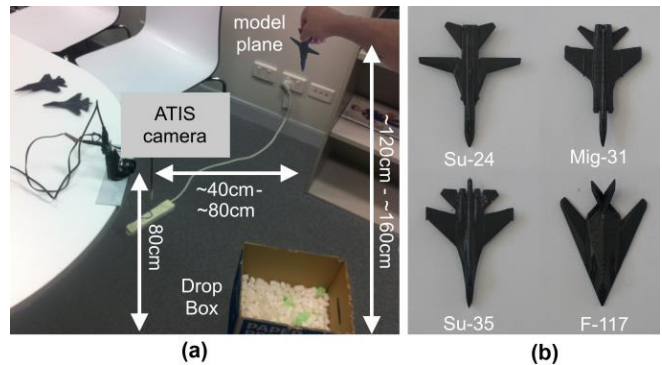

**Figure 1: Data collection setup for the airplane dropping dataset.** (a) The physical setup used for recording dataset in which an ATIS camera is attached to a table and the airplanes dropped freehand in front of the camera. (b) A top-down and labelled view of the four model airplanes used to generate the dataset.

The models were dropped 100 times each from a distance ranging from 120 cm to 160 cm above the ground and at a horizontal distance of 40 cm to 80 cm from the camera. This ensured that the airplanes passed rapidly through the field of view of the camera, with the planes crossing the field of view in an average of 242 ms ± 21 ms. No mechanisms were used to enforce consistency of the airplane drops, resulting in a wide range of velocities and entry points into the field of view of the camera. Additionally, there were variable delays before and after each drop, resulting in recordings of varying lengths. The dataset was additionally augmented with left-right flipped versions of the recordings resulting in 200 drops for each airplane type.

The variations in airplane drops create a more realistic dataset and pose a significantly more challenging task for the tracking and classification algorithm. An example of the variability in the airplane drops is demonstrated in Figure 2, which shows binned events in the same 3 ms slice of data from 20 randomly selected recordings from the dataset. It is apparent from the figure that there are significant variations in the positions of the airplanes, their orientations and their sizes.

The recordings were captured using the same model of ATIS camera and the same acquisition software used in capturing the N-MNIST dataset in [12], and the recordings were stored in the same file formats, thereby maximizing compatibility with other neuromorphic algorithms and systems.

The generated dataset is constrained in the sense of having a single high-speed object in the field of view against an effectively blank background. This restriction allows an efficient focused investigation of different methodologies as well as of the sources of variance in the data such as target orientation and velocity. This restriction also limits the generalization of the results to more complex scenes. Instead, the dataset and resulting solutions can more usefully be viewed as describing localized information and processing in a small region of a larger more complex visual scene. The full dataset can be found at [16].

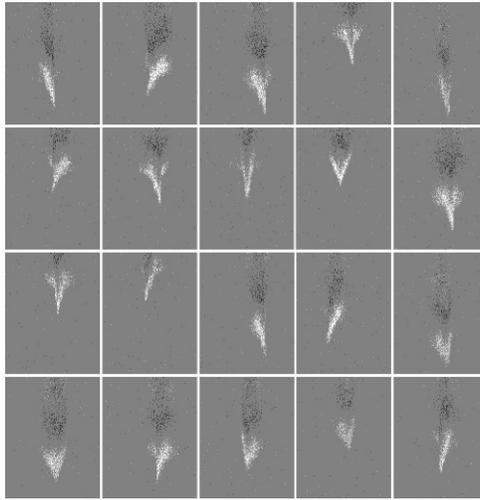

**Figure 2: Examples of the variation in the dataset in terms of position, scale, orientation and speed.** Each image represents a frame rendered from the same 3 ms of events extracted from each recording with ON events represented with white pixels and OFF events represented with black pixels. The twenty random samples clearly demonstrate the difficulty of the recognition task. Airplane class key ordered from top left to bottom right, Mig-31: {2, 3, 7, 11, 12}, F-117: {9, 15, 16, 18, 19}, Su-24: {1, 5, 8, 14, 20} and Su-35: {4, 6, 10, 13, 17}.

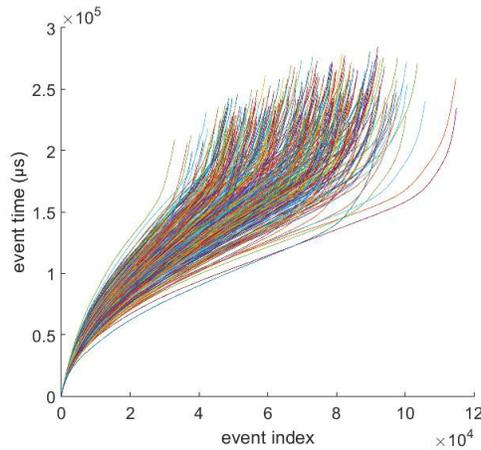

**Figure 3. Event timestamp profiles of all airplane drops in the dataset.** The above plot shows the timestamp of each recording as a function of the index of each event. Each curve represents the timestamp profile for an individual recording in the dataset and demonstrate the variable rates of event generation over time and across different airplane drop recordings. These differences are a function of the speed, size, and shape of the airplanes and the distance from the camera.

*B. Time-Based vs Event-based Decaying Memory Surfaces*

An event $ev_i$ from the ATIS camera can be described mathematically by:

$$ev_i = [\mathbf{x_i}, t_i, p_i]^T \quad (1)$$

where $i$ is the index of the event, $\mathbf{x_i} = [x_i, y_i]$, is the spatial address of the source pixel corresponding to the physical location on the sensor, $p_i \in \{-1, 1\}$ is the polarity of event indicating whether the log intensity increased or decreased, and $t_i$ is the absolute time at which the event occurred [17]. The time $t_i$ is applied to the event by the ATIS camera hardware and has a resolution of one microsecond.

Event-based algorithms require iterative processing of each event, and therefore require that each new observation be combined with previously observed local events, both in space and in time. This is accomplished using a variation of the time surfaces from the HOTS algorithm [11], but extended to cover both surfaces based on time and based on event index. Each new incoming event updates the surface and defines a region representing the spatio-temporal neighborhood on which further processing may be performed. As these surfaces encompass both time and index, we will refer to them as memory surfaces from this point onward.

The timing and polarity information contained in each event, as shown in equation (1), allows the generation of two useful surfaces, based on time and polarity, from which more complex surfaces can be constructed. The first surface, referred to as $T_i$, maps the time of the most recent event to spatial pixel location and is described in (2), with the corresponding surface $P_i$ for event polarity given by (3). Note that in this work we only used events with positive polarity, i.e., $P_i = 1$.

$$T_i : \mathbb{R}^2 \rightarrow \mathbb{R}$$

$$\mathbf{x}: t \rightarrow T_i(\mathbf{x}) \qquad (2)$$

$$P_i: \mathbb{R} \rightarrow \{-1, 1\}$$
$$\mathbf{x}: p \rightarrow P_i(\mathbf{x}) \qquad (3)$$

We compare two approaches for memory surface decay. In the HOTS algorithm, making use of similar functions and surfaces, the memory of recent events is decayed as a function of time. A second approach, explored here, decays the events not as a function of time, but in response to new incoming events. We then define the analogous function to (2) for index-based surfaces. This surface, $I_i$, is defined in (4) and stores the indices of incoming event for each spatial pixel.

$$I_i: \mathbb{R}^2 \rightarrow \mathbb{R}$$
$$\mathbf{x}: i \rightarrow I_i(\mathbf{x}) \qquad (4)$$

In addition to exploring time-based decay and index-based decay, three different transfer functions or memory kernels are investigated. These kernels are event binning (*BTS/BIS*), linear decay (*LTS/LIS*) and exponential decay (*ETS/EIS*). As a point of reference, the HOTS algorithm makes use of exponential decaying time kernels.

In all surface generation methods, when a new event arrives, the memory surface at $x_i$ is set to $P_i$. When using the event binning technique, the value on the surface maintains its value over a temporal window $\tau_e$ or index window $N_e$ after which it is reset to zero. The event binning method for surface generation is described by equations (5) for the time-based binning (*BTS*) and (6) for the event-based binning (*BIS*).

$$BTS_i(\mathbf{x}, t) = \begin{cases} P_i(\mathbf{x}), & T_i(\mathbf{x}) - t \leq \tau_e \\ 0, & T_i(\mathbf{x}) - t > \tau_e \end{cases} \qquad (5)$$

$$BIS_i(\mathbf{x}) = \begin{cases} P_i(\mathbf{x}), & I_i(\mathbf{x}) - i \leq N_e \\ 0, & I_i(\mathbf{x}) - i > N_e \end{cases} \qquad (6)$$

For the linearly decaying time surface (*LTS*) and linearly decaying index surface (*LIS*) the initial value set on the surface in response to a new event instead decays toward zero linearly as a function of time. These surfaces are described by (7) for time-based linear decay or in response to incoming events as described by (8) for event-based linear decay.

$$LTS_i(\mathbf{x}, t) = \begin{cases} P_i(\mathbf{x}) \cdot (1 + \frac{T_i(\mathbf{x}) - t}{2\tau_e}), & T_i(\mathbf{x}) - t \geq 2\tau_e \\ 0, & T_i(\mathbf{x}) - t < 2\tau_e \end{cases} \qquad (7)$$

$$LIS_i(\mathbf{x}) = \begin{cases} P_i(\mathbf{x}) \cdot (1 + \frac{I_i(\mathbf{x}) - i}{2N_e}), & I_i(\mathbf{x}) - i \geq 2N_e \\ 0, & I_i(\mathbf{x}) - i < 2N_e \end{cases} \qquad (8)$$

The exponential decay method works in a similar manner to the linear decay, with the value placed on the surface decaying exponentially instead of linearly with respect to either time or event. This results in the equations for the exponentially decaying time surface (*ETS*) shown in (9), and the exponentially decaying index surface (*EIS*) shown in (10).

$$ETS_i(\mathbf{x}, t) = P_i(\mathbf{x}) \cdot e^{\frac{T_i(\mathbf{x}) - t}{\tau_e}} \qquad (9)$$

$$EIS_i(\mathbf{x}) = P_i(\mathbf{x}) \cdot e^{\frac{I_i(\mathbf{x}) - i}{N_e}} \qquad (10)$$

The equations for these surfaces make use of a constant parameter, time constant $\tau_e$ for time-based methods and index constant $N_e$ for the index-based methods and the chosen values for these parameters are demonstrated in Figure 4(a) and (b). These graphs show the integration of the kernels for the memory surfaces over a period of 3 ms in (a), and over 554 events in (b). These values were chosen based on the mean data rate over all recordings.

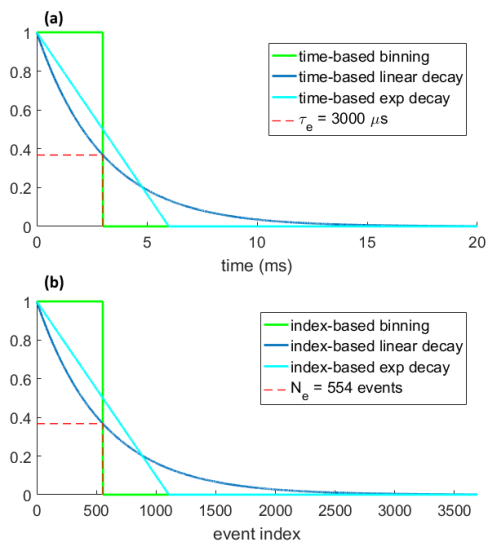

**Figure 4. Plots of the six methods for generating time and index surfaces.** Panel (a) Shows the three time-based kernels over time. Note that the area under all kernels is the time constant $\tau_e = 3$ ms. Panel (b) shows the value of the event-based kernel as a function of event index. Here the mean dataset event rate over all recordings (~184.5k events/s) was used to obtain equivalent sized kernels with index constant $N_e = 554$ events.

Figure 4 demonstrates that the integral under each of these kernels is equal given the event rate. Therefore, the surface generation methods will generate equivalent total surface activation over the entire dataset.

Figure 5 illustrates the difference between time-based and the index-based decaying surfaces for a single recording from the dataset. The figure demonstrates the relative stability of total surface activation when using index-based surfaces. The figure shows that the binning time surface has a lower activation than the binning index surface when the speed of the airplane is low (at the start of the recording). As the airplane speeds up, the total time surface activation continues to increase whilst the index surface remains approximately constant. In fact, at t = 157 ms, the total activation on the time surface is approximately twice that of the index-based surface. This demonstrates the speed-invariant properties of the index-based surfaces.

The effects are most noticeable when using the binning time surfaces, but are also present on the exponential and linear decaying surfaces. **Figure 6** shows the total surface activation of every recording in the dataset superimposed on one another for the various types of surface. The mean activation is shown as a dashed black line. **Figure 6** highlights the significantly different profiles resulting from the index-based approach.

It is also worth noting that for the time-based decay surfaces, the exponentially decaying kernels generate noticeably lower surface activations than linearly decaying kernels, which in turn generate lower activations than binning. This drop is due to the arrival of new events overwriting the entries for pixels that have recently been activated. This effect is more pronounced for kernels with a longer time window as the surface maintains the value for longer. This same effect is also present in the index-based decay plots but is less prominent due to the lower variance of the index-based activation plots.

### C. Target Velocity vs Surface Activation

Prior to the feature extraction and recognition, the airplane is detected and the location within the field of view is determined. The speed of the airplanes is much faster than any other stimulus expected within the field of view of the camera, such as the body of the author accidently entering the frame as can be seen in the lower right pane of Figure 7d. Therefore, summation of events across the rows and columns of the camera's field of view (after normalization and thresholding as shown in Figure 7b and c) provides a simple and effective method to detect the boundary of the airplane reliably, even in the presence of other (relatively slow) moving distractors.

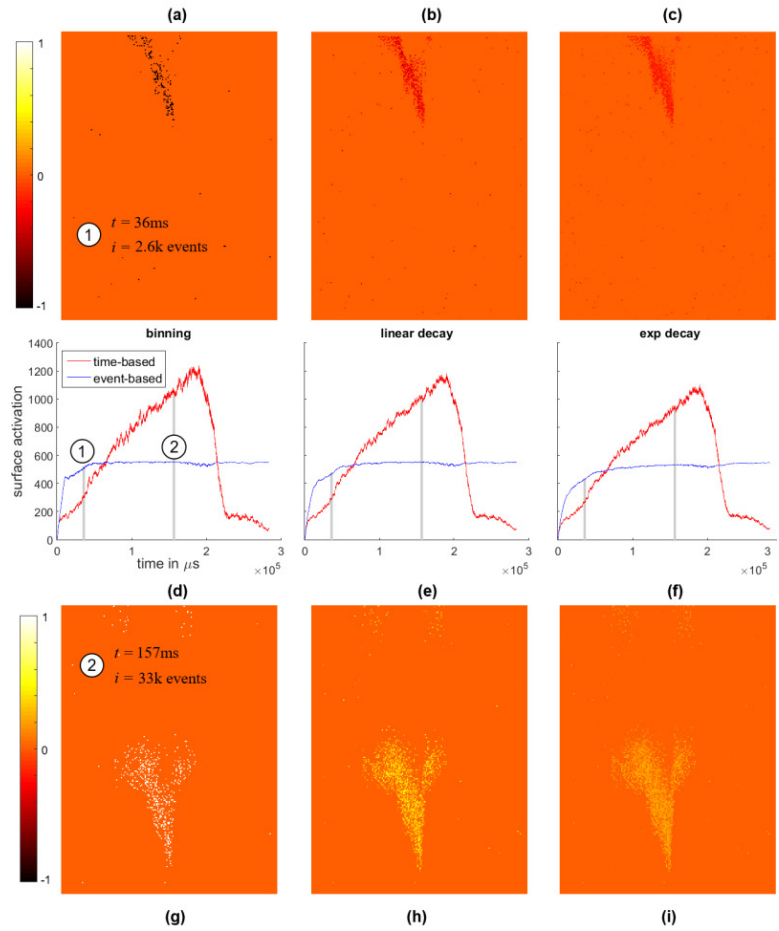

**Figure 5. Comparison of surface activation for a single recording**. Panels (a), (b), and (c) show the surface differences $(BTS_i - BIS_i)$, $(LTS_i - LIS_i)$ and $(ETS_i - EIS_i)$ respectively at the beginning of the recording ($t = 36$ ms). This moment in the recording is marked (1) on panel (d) which displays total surface activation for the binning method $\sum_{x,y} BTS_i$ and $\sum_{x,y} BIS_i$. The two traces in panel (d) show that at the beginning of the recording when the target airplane's speed is low the binning time surface has a lower activation than the binning index surface. However, as the target speeds up, the total time surface activation also increases, while the index surface remains approximately stable, such that by $t = 157$ ms the time surface activation $\sum_{x,y} BTS_i$ is approximately twice that of $\sum_{x,y} BIS_i$. Panels (e) and (f) show a similar but slightly less pronounced relative increase for the linear and exponential decay surfaces. Panels (g), (h), and (i) show this relative increase for the binning, linear and exponential decay surfaces by plotting the differences of (a) – (c) at $t = 157$ ms.

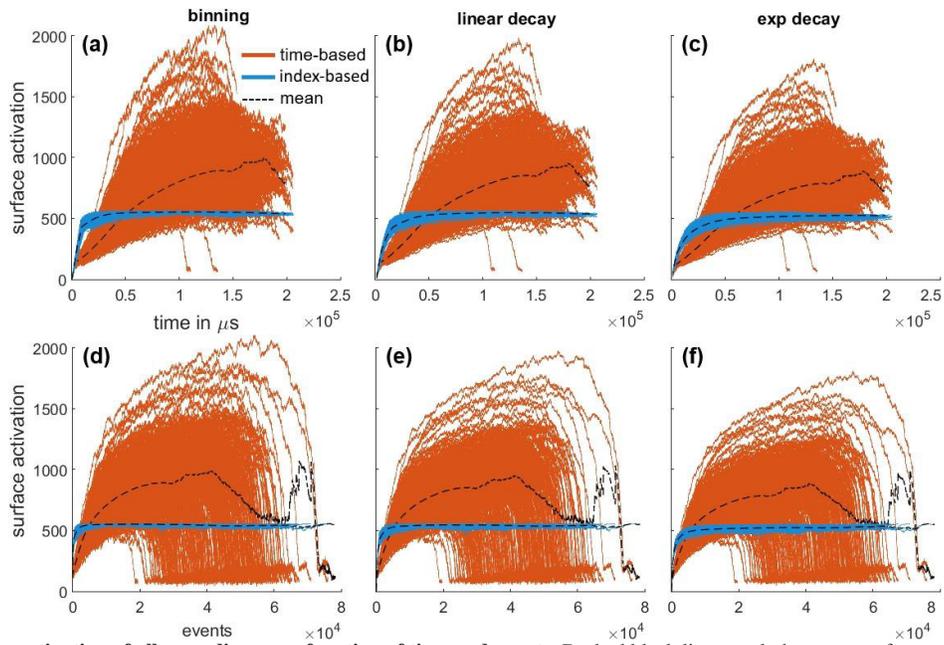

**Figure 6. Total surface activation of all recordings as a function of time and events.** Dashed black lines mark the mean surface activation for each of the time-based and index-based decay methods.

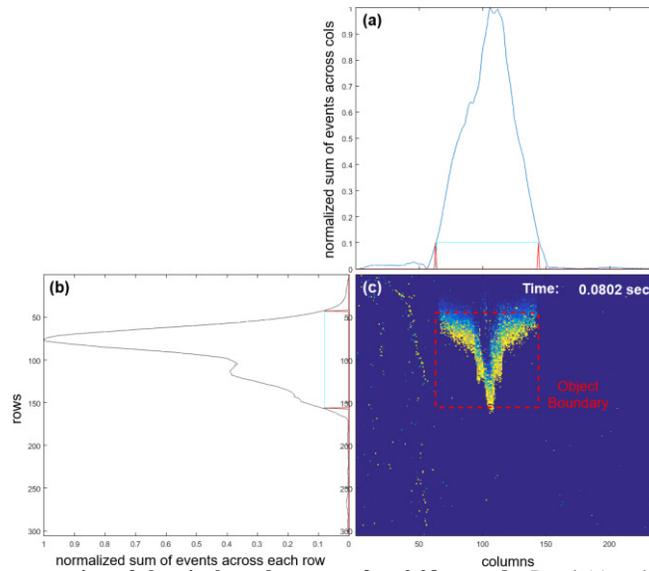

**Figure 7. Screenshot from a live demonstration of the airplane drop test after 0.08 seconds.** Panel (a) and (b) are a smoothed summation of recent events across columns and rows respectively. The smoothing was performed by using an 8-pixel wide rectangular moving average window. Due to the relatively high speed of the airplane these summations, when normalized and thresholded at 0.1, could reliably be used to extract the fast-moving airplane from the static background or slower moving objects. The generated target object's boundary is shown in (c. Note that movement of the body of the author (light vertical trace on the left) as he drops the airplane is slow relative to the airplane and generates relatively few events and so does not reach even the low set detection threshold level.

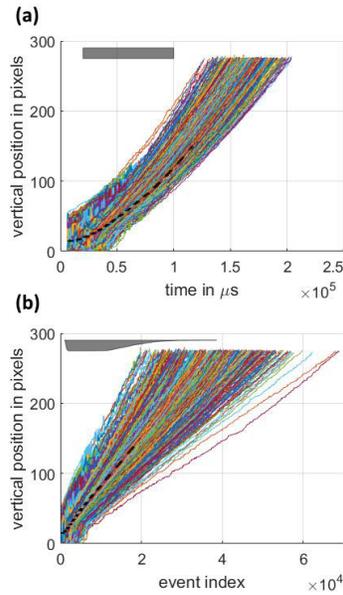

**Figure 8. Estimated vertical position of the target as a function of time (a) and as a function of event index (b).** The dashed black line marks the mean position over all recordings. For the entire dataset, the mean time interval from the first valid object boundary detection event to the last was 156.2 ms seconds with a standard deviation of 17.8 ms. The target's position was defined as the midpoint between the object boundaries as shown in Figure 7 (d). The grey bar at the top left in (a) indicates the time window used for investigating the effect of target velocities on surface activation in Figure 9. The same grey time window bar is shown in lower (b) panel as a function of event index. The relative thickness of the bar is proportional number of recordings in the time window of (a) at each event index.

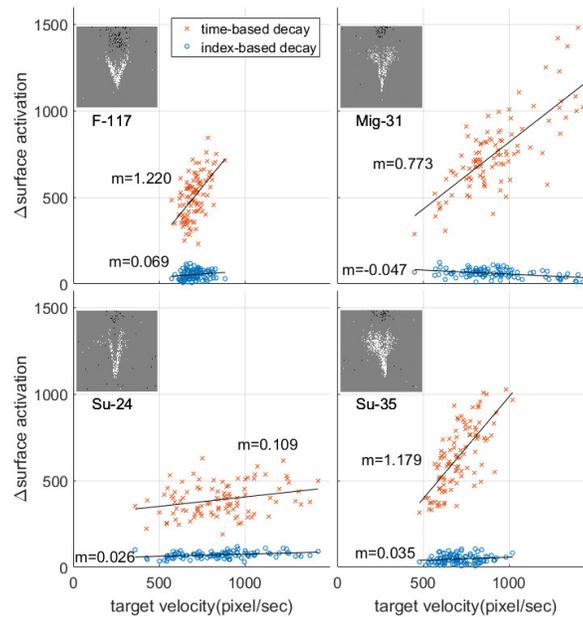

Figure 9. Relationship between change in surface activation and target velocity for each airplane class in the dataset and the resultant mean rate of change in surface activation over the time window indicated in **Figure 8**. For each plot m indicates the slope of the line of best fit.

This method is used to detect and calculate the vertical position of each airplane drop recording, and the estimated vertical position of each target airplane is shown in Figure 8, both in terms of time, Figure 8(a) and event index Figure 8(b). These vertical position profiles serve to further highlight the difference between the index-based and time-based approaches. Whereas the estimated position over time take on their expected parabolic shape, when the vertical position is plotted against index, the trajectories are linear to the first approximation.

Figure 9 illustrates the wide range of velocities in the data set and the associated mean rate of change in surface activation for index-based and time-based decaying surfaces using exponential decay kernels. The line of best fit through the data demonstrates different relationships between velocity and change in surface activation which arise from the different geometries of the airplanes. In all cases, however, surface activation is significantly more sensitive to velocity when using time-based surfaces.

## D. Event-based Feature Extraction

An event-based feature extractor was used to learn the most common spatio-temporal features generated by the recordings. We used an unsupervised spike-based feature extraction algorithm developed for hardware implementation, as previously described in [18]. In this algorithm, the Synapto-dendritic Kernel Adaptation Network (SKAN), a single layer of neurons with adaptive synaptic kernels and adaptive thresholds compete in the temporal domain to learn commonly observed spatio-temporal spike patterns. These adaptive synapto-dendritic kernels provide an abstracted representation of the coupling of pre- and post-synaptic neurons via multiple synaptic and dendritic pathways allowing unsupervised learning and inference of precise spike timings. In [19] the algorithm was extended using a simplified model of Spike Timing Dependent Plasticity [20] to provide synaptic encoding of afferent SNR. In [21] the algorithm was used to perform real-time unsupervised hand gesture recognition using an FPGA. In this work, the event-based approach is continued at the feature extraction layer wtih the output spike of the winning neuron representing a *feature event*. When the camera detects a new event, a 13×13-pixel region from the event memory surface around it is converted to a temporally coded spatio-temporal spike pattern. The normalized real-valued intensity of the surface is first rescaled from 0-1 to 0-255 and then mapped to an 8-bit unsigned integer. This integer representation of the local surface region is then encoded into spike delays forming a spatio-temporal spike pattern. The resultant pattern is then used as the input to a 25-neuron network. The neurons were trained using 50 recordings from each plane type augmented by the left-right flipped version of these recordings. After presentation of these 400 training recordings the encoded features were found to have stabilized. Learning (adaptation) in the feature detection neurons was then disabled. The resultant spatio-temporal features learnt by SKAN are shown in Figure 10.

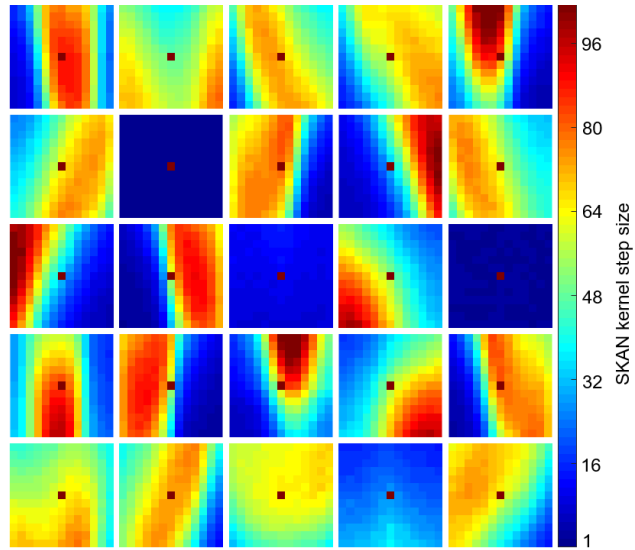

Figure 10. The twenty-five spatio-temporal features extracted from the ATIS airplane drop dataset. Specialized features coding for the leading edge of airplane nose cone and wings as well as variants of the noise spike often produced by the ATIS camera dominate the feature set resulting in a well separated set of feature inputs to the classifier. Only the feature-set obtained from the time-based, exponentially decaying surface is shown above for brevity. The features resulting from the other memory surfaces are similar.

Following feature extraction, and with learning disabled, the neurons compete to recognize incoming spatio-temporal event patterns generated from the same 13×13-pixel region of the event memory surface following each new event with the spike output of the winning neuron representing a feature event. These feature events were then stored onto 25 separate *feature memory surfaces*, which were generated identically to the event surfaces described in section II B.

## E. Spatial Pooling of Feature Surfaces

In order to reduce the required processing and speed up simulation, the subsystems following the feature memory surfaces were operated in a frame-based manner such that at periodic intervals the estimated target region from each feature surface was sampled to generate feature frames. The interval used for sampling was the same as the time surface decay constant $\tau_e$ = 3 ms. To perform spatial pooling, the estimated object boundary region was summed along the rows and columns, generating two one dimensional feature vectors (histogram of feature events), one for the rows and one for the columns. The length of these vectors would vary at each feature frame depending on the size of the estimated target region. In order to provide the classifier with a uniform input layer size, the varied length feature vectors were therefore resampled, using linear interpolation, to a uniform vector size of 72, which when multiplied by the number of pooling dimensions (2), and the number of features (25), produced a 3600-input layer for the classifier. The feature extraction and feature memory surface generation and pooling operations represented the bulk of the processing time and were performed only once for each of the decay functions when collating results. Early testing showed the same features were generated through multiple simulations of the feature extraction network and different random samplings of the dataset. The resultant end-to-end system is shown in Figure 11.

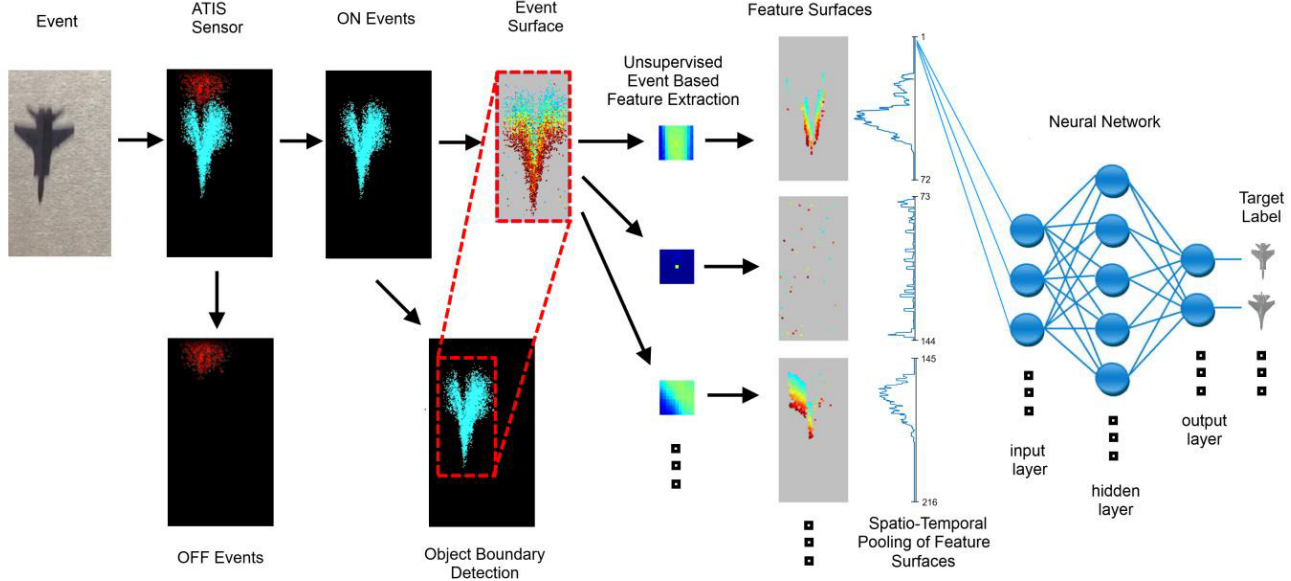

Figure 11. Block diagram of the full event-based tracking feature extraction and recognition system.

*F. Parameter Selection*

In order to fairly evaluate the relative performance in terms of recognition accuracy resulting from different decay kernels, surfaces decay methods, feature extractor numbers and their receptive field sizes, a large number of free system parameters must first be selected. These parameters, listed in Table 1, are used to implement event and feature surface generation, surface sampling, object detection and tracking, feature extraction and recognition, spatial pooling, and classification and regularization. In order to ensure that the selected parameters do not advantage the index-surfaces or the feature extraction method used in this work, all subsystem parameters would need to be evaluated in terms of their combined effects on the performance of each method under testing. However, this represents a prohibitively large search space to explore in a brute force fashion. Instead, the approach taken in this work to remove possible parameter selection bias in favor of the proposed methods was to optimize all parameters to achieve the highest recognition accuracy on what may be considered the null hypothesis: that simple time-based binning kernels used on raw input events outperform other kernels, decay methods, and feature extractors. To this end the parameters in Table 1 and all algorithm design choices where found via a manual heuristic search for optimal recognition performance using the time-based binning surface $BTS_i$ whose spatially pooled output was fed directly to the classifier without the use of feature extractors. The classifiers were then selected and precisely regularized for optimal performance on the output data generated by the selected parameters. Once optimized in this way for the 'null hypothesis', the same parameters and network structure were used for all other tests, ensuring that recognition results were biased in favor of the simple time-based binning approach and not those proposed in this work.

| Subsystem - Parameter | value |
|---|---|
| Surface generation - Time constant | $\tau_e$ = 3ms |
| Surface generation - Index constant | $I_e$ = 554 events |
| Tracker - Smoothing window size | 8m pixels |
| Tracker - Smoothing window type | Moving average |
| Tracker - Normalized threshold | 0.1 |
| SKAN - Number of features | 25 |
| SKAN - Number of input channels | 13×13 = 169 |
| SKAN - Other parameters | Same as [18] |
| Classifier - Input size using raw event surface (E) | 72×2 = 144 |
| Classifier - Input size feature event surfaces (F) | 72×25 = 1800 |
| Classifier - ELM hidden layer size | 30,000 Neurons |
| Classifier - Surface sampling interval | 3ms |

Table 1. Free parameters used in the system (unless otherwise stated).

*G. Classification*

*1) Choosing classifiers*

To evaluate the performance of the system, two measures of recognition accuracy were considered: per-frame accuracy and per-drop accuracy. For the per-frame measure, the feature vectors described Section II.D were presented to the classifier at periodic time intervals $\tau_e$. At each frame, the class with the largest output was selected as the winner for that frame. For the per-drop accuracy measure the class with the highest number of per-frame wins was selected.

A linear classifier and an Extreme Learning Machine (ELM) classifier [17] with a hidden layer size of 30000 neurons was trained using the time-based binning method to achieve the highest per-frame recognition accuracy. Figure 12 details the results from this parameter search and the selected classifiers.

## III. RESULTS

### A. Results on Full Dataset

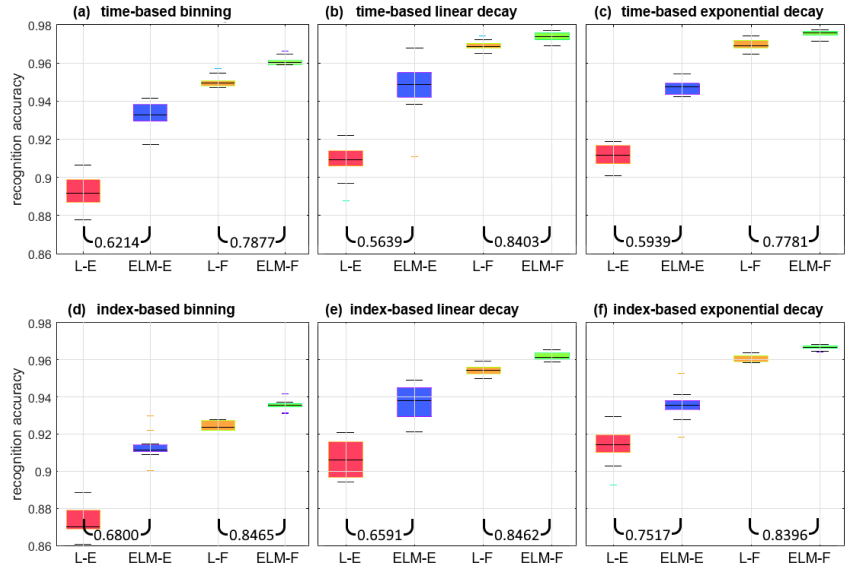

Figure 12. Per-frame recognition accuracy on the full dataset over n = 20 independent trials. Each panel shows results from four network arrangements. In (L-E), and (ELM-E) the linear classifier and 30K hidden layer ELM chosen in section II.G.1) operate on inputs from raw event memory surfaces respectively. In (L-F), and (ELM-F) the same classifiers use 25 feature surfaces as inputs. Each panel shows results for a different surface generation method: The top three panels show time-based methods using (a) binning, (b) linear and (c) exponentially decaying surfaces. The bottom three panels show corresponding index-based binning (d), linear (e) and exponentially decaying surfaces(f). The two ratios at the bottom of each panel indicate the median error ratio of the ELM over the linear classifier.

The per-frame recognition results on full dataset are shown in Figure 12. For each of the panels, the same performance pattern is observed: When operating raw event memory surfaces as inputs, the large capacity ELM (ELM-E) significantly outperforms the linear classifier (L-E). This demonstrates the non-linearity of the classification boundaries in this case. In comparison, when feature surfaces are used as inputs, the improvement margin gained by the ELM (ELM-F) is small relative to the linear classifier (L-F) suggesting that the output of the 25 feature extractors is significantly more linearly separable with less room for improvement through further non-linear processing. Also noteworthy is that the linear classifier operating on feature surfaces (L-F) outperforms the ELM operating on the event surfaces (ELM-E) for all surfaces generation methods. This shows that the application of a small number of trained local feature extractors is more effective than using a much larger globally connected network of neurons with random input weights.

Comparing the results across the panels, the exponentially decaying surfaces outperform linear surfaces, which in turn outperform binning surfaces. The rigorousness of this result is further bolstered by the fact that the system parameters were optimized for the time-based binning method. These results confirm the suitability of exponential kernels for time and index-surface generation. This conclusion is also supported by results in [22], where the information from the visual scene is found to rapidly rise within a small initial temporal window but thereafter fall gradually with increasing window size. By weighing events in an approximately compensatory manner to their information content as described in [22], the exponentially decaying kernel results in the highest information content for the classifier.

Another observation from Figure 12 is that all time-based decay methods outperform the index-based decay methods on the full dataset with the largest performance disparity observed between the time-based binning $BTS_i$ and index-based binning methods $BIS_i$.

Based on the results shown in Figure 12 we narrow further investigations by selecting linear classifiers L-E and L-F and focus on exponentially decaying surfaces $EIS_i$ and $ETS_i$.

### B. Frame Balanced Dataset

In order to generate a balanced dataset an equal number of frames from each recording was selected. In this way, the total number of presentations to the classifier for each class was equalized. As Figure 13 shows 1, 2, 4, 8, 16 and 32 frames were sampled from each of the airplane recordings and presented to the linear classifier operating on events surfaces L-E and feature surfaces L-F for each of the $EIS_i$ and $ETS_i$ surfaces.

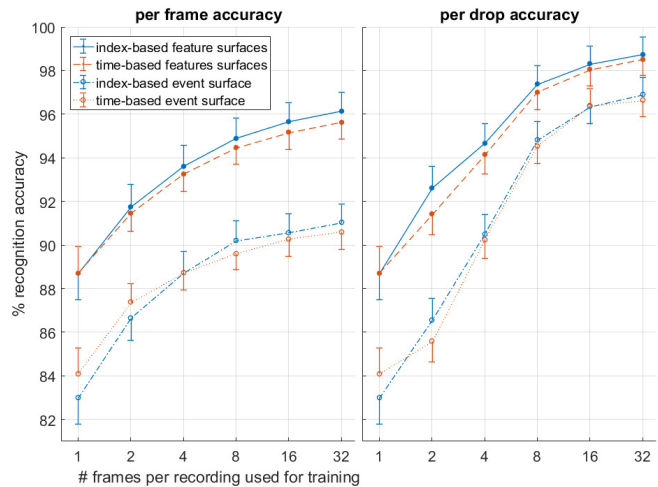

Figure 13. Comparison of per-drop and per-frame recognition accuracy as a function of the number of randomly selected frames used during training from each recording. The index-based $EIS_i$ surface and time-based $ETS_i$ surfaces are compared. Over N = 20 trials using linear classifier the index-based and time-based surfaces was observed.

As Figure 13 demonstrates that both the per-frame and per-drop accuracy increase as a function of the number of frames used during training. Additionally a sharper increase and higher final accuracy is observed for the per-drop accuracy measure as would be expected. The relative performance margin of the network utilizing feature surfaces over the raw event surfaces is reduced in the per-drop measure as more information is accumulated over a recording, reducing error and approaching the 100% accuracy upper bound. The highest number of random frames used per recording was 32 as this was the total number of frames in the shortest recording. Table 2 details the accuracy results for this balanced dataset while Figure 14 shows misclassified recordings for one instance of the highest performing network using index-based decaying feature surfaces and a linear classifier. They illustrate that some drops are impossible to classify correctly.

Interestingly in contrast to the full dataset results detailed in section III.A, the frame balanced results in Figure 13 and Table 2 show little significant difference in accuracy between the index-based and time-based surfaces for either the per-frame or per-drop measures suggesting that the observed slight advantages in accuracy on the full dataset may be due to the use of time-based surfaces during parameter selection of section II.F and linked to imbalances in the number of frames per recording present in the full dataset.

|  | Per-frame | Per-drop |
|---|---|---|
| Time-based Event surface | 90.60 +/-1.02% | 96.64 +/-1.47% |
| Index-based Event surface | 91.03 +/-0.89% | 96.90 +/-1.34% |
| Time-based Feature surfaces | 95.64 +/-0.79% | 98.52 +/-0.75% |
| Index-based feature surfaces | 96.15 +/-0.84% | 98.75 +/-0.78% |

Table 2. Per-frame and Per-drop accuracy results on the frame balanced dataset for four selected systems: Linear classifier operating on events surfaces (L-E) and feature surfaces (L-F) for each of the $EIS_i$ and $ETS_i$ surfaces. Number of trials used 20.

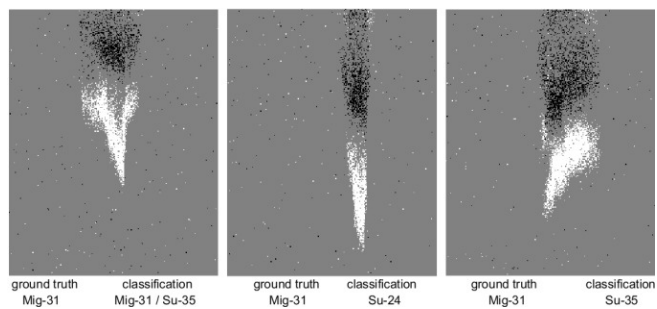

Figure 14. The three miss-classified drops by an instance of a linear classifier using 25 exponentially decaying index-based feature surfaces. Captured frames show airplanes at mid-point (in time) of recording.

## C. Velocity Segregated Dataset

In order to compare index-based and time-based surfaces in terms of target velocity invariance, the recordings were divided into two hundred 'slow' and two hundred 'fast' recordings based on the estimated vertical airplane velocity at the midpoint (in time) of each recording. Since the airplanes speed up during the fall, the system was trained on the n-first (slowest) frames of the slow recordings and tested on the n-last (fastest) frames of the fast recordings. In this way by varying the number of frames n, datasets with different degrees of velocity segregation could be tested. The resulting recognition accuracies in Figure 15 demonstrate that with increasing n, and decreasing velocity segregation in the data, the recognition accuracy of all systems rise. Figure 15 further shows that although training on a speed segregated dataset significantly reduces accuracies for all systems in comparison to training using a randomly sampled dataset such as shown in Figure 13, the decline is significantly sharper for time-based decaying surfaces. This difference demonstrates the relative robustness of index-based decay surfaces to variance in velocity and their utility in applications where the full range of potential target velocities to be encountered during testing is not available in the training data.

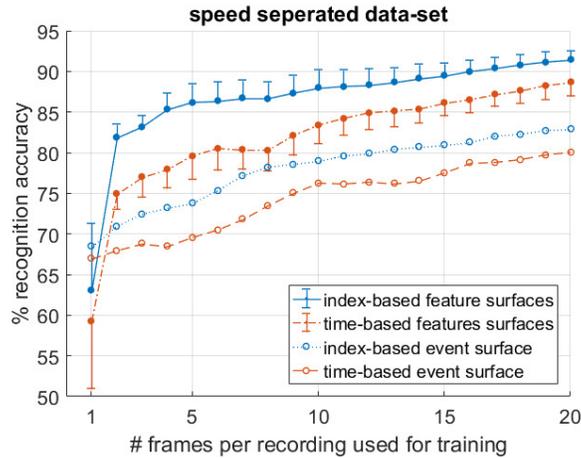

Figure 15. Mean and standard deviation per-frame accuracy on a speed segregated dataset over ten trials clearly demonstrates superior performance of index-based surfaces in the presence of velocity varying data.

Therefore, given the results in the previous section, it can be concluded that systems using index-based decay surfaces tend to match equivalent systems using time-based decay surfaces when presented with an adequately wide range of velocities in the training data since their advantage of velocity invariance is effectively neutralized. But when the available range of velocity distributions is incomplete, index-based decay surfaces tend to produce better results. Given this finding, in the next section we narrow our focus exclusively on index-based surfaces and investigate the effect of different feature extraction networks and their effect on recognition accuracy.

## D. Feature Extractors

In order to characterize the effectiveness of the feature extraction subsystem in an unbiased manner, a range of feature sizes and number of feature extractors were investigated and assessed in terms of the resultant recognition accuracy. In addition, for each point on the feature size-feature number space, the results of the learning algorithm described in II.D was compared to those of equivalent sized networks with random feature sets. The mean accuracy results in Figure 16 (top panels) demonstrate that learnt features outperform random features at every scale while exhibiting slightly lower variance in accuracy (bottom panels).

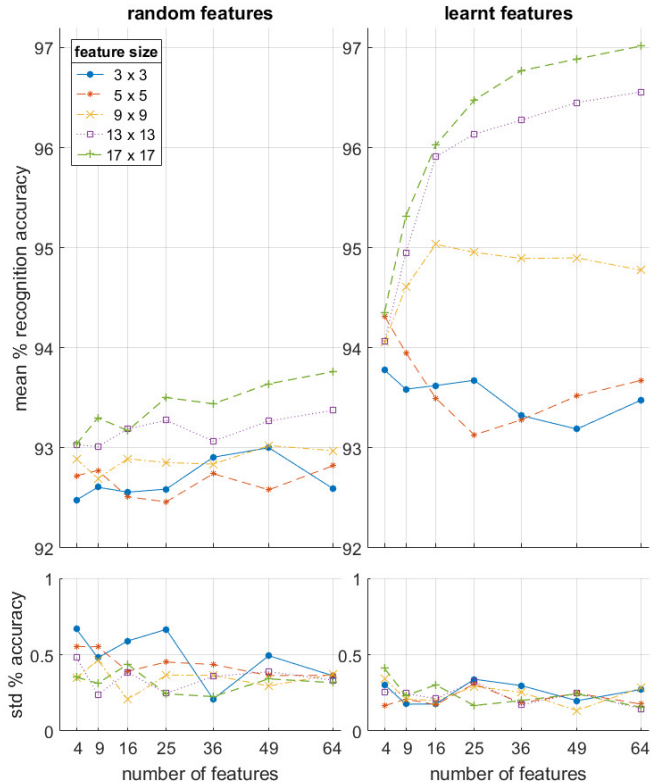

Figure 16. Per-frame accuracy on the full dataset as a function of feature size and number of features used in the feature extraction layer for both learnt and random features. N = 10 independent feature sets with 10 cross validating classifications per feature set. Note that the baseline linear accuracy using the raw event surface with no feature extraction layer was 91.38 +/- 0.81%.

In addition, while the results from the random features suggest a slight trend toward increased accuracy as a function of both feature numbers and feature size, the learnt feature results clearly show that the larger feature sizes (17x17 and 13x13) generate higher accuracy as the number of features increases, while smallest feature sizes (3x3 and 5x5) exhibits a weak downward trend with number of features. When the feature size is small, only a few distinct combinations exist. Therefore, when and a large number of them are used, several features will be very similar, resulting in near identical input generating very different input to the classifier. The accuracy results of the mid-sized features (9x9) initially rise with feature numbers before declining at twenty five and more features. For both the random and learnt feature sets, the feature size has little effect on accuracy when the number of features becomes very small.

IV. DISCUSSION

In this dataset, the recordings were varied systematically to cover a wide range of target speeds. As such any random splitting of training and testing data provided an overlapping range of target speeds in both set. This overlap removed the need for and advantage of index-based decaying surfaces which provide robustness to target velocity. However, in many applications such an overlap may be missing. This condition was simulated here by segregating the data-based on speed, highlighting the utility of the index-based decay method in such a data-limited context.

The central weakness of the index-based decaying method is that it can only be used locally on a single target. If events from other non-target object cause a decay in surface activation of the target, vital information would be lost. However this limitation can be overcome if target segregation is already assumed to have occurred via an upstream system. Equivalently, the presented dataset and the resulting performance of the index-based systems can be viewed as focusing on a locally operating subsystem within a larger processing system.

The linearly decaying memory surfaces examined in this work were shown to perform nearly as well as exponentially decaying surfaces in a challenging tracking and recognition task. While using linearly decaying surfaces allows slightly faster computation it can potentially allow significantly more efficient implementations of event surfaces in neuromorphic hardware relative to exponentially decaying surfaces.

In order to rigorously investigate the effects of different kernels, decaying methods, feature sizes and numbers we limited the exploration to a single layer network. Yet the design of deeper networks can be informed by these single layer results.

## V. CONCLUSION

In this work we investigated the use of event-based unsupervised feature extractors together with time decaying event and feature memory surfaces for use in tracking, feature extraction, and classification. Using a dataset featuring a range of target shapes, scales, orientations, and velocities it was observed that exponentially decaying kernels outperform other kernel and that index-based decaying surfaces perform equally as well as time surfaces and when robustness to target speed is not required, and outperform them when it is required. We also showed a clear superiority of learnt features over random features and the largest networks of neurons with the largest receptive fields outperformed all other configurations.